\documentclass[a4paper,conference]{IEEEtran}

\def\BibTeX{{\rm B\kern-.05em{\sc i\kern-.025em b}\kern-.08emT\kern-.1667em\lower.7ex\hbox{E}\kern-.125emX}}

\usepackage[utf8]{inputenc} 
\usepackage[T1]{fontenc}    
\usepackage{times}  
\usepackage{helvet} 
\usepackage{courier}  
\usepackage[hyphens]{url}  
\usepackage{graphicx} 
\usepackage{natbib}  
\usepackage{caption} 
\usepackage{booktabs}       
\usepackage{nicefrac}       
\usepackage{microtype}      
\usepackage{caption}
\usepackage{listings}
\usepackage{multirow}
\lstdefinestyle{mystyle}{
        language=python,
        basicstyle=\footnotesize\ttfamily,
        breaklines=true,
        keepspaces=true,
    }
\usepackage{adjustbox}
\usepackage{sidecap}
\usepackage{subcaption}
\usepackage{makecell}
\usepackage{amsmath}
\usepackage{lineno}
\usepackage{enumitem}
\usepackage{comment}
\usepackage{array}
\newcolumntype{L}{>{\centering\arraybackslash}m{3cm}}

\ifCLASSINFOpdf
\else
\fi

\begin{document}
%
\title{Biomedical Named Entity Recognition at Scale}

\author{\IEEEauthorblockN{Veysel~Kocaman}
\IEEEauthorblockA{John Snow Labs Inc.\\ 
  16192 Coastal Highway\\
  Lewes, DE , USA 19958\\
veysel@johnsnowlabs.com}
\and
\IEEEauthorblockN{David Talby}
\IEEEauthorblockA{John Snow Labs Inc.\\ 
  16192 Coastal Highway\\
  Lewes, DE , USA 19958\\
david@johnsnowlabs.com}
}


%


\maketitle

\begin{abstract}
Named entity recognition (NER) is a widely applicable natural language processing task and building block of question answering, topic modeling, information retrieval, etc. 
In the medical domain, NER plays a crucial role by extracting meaningful chunks from clinical notes and reports, which are then fed to downstream tasks like assertion status detection, entity resolution, relation extraction, and de-identification. 
Reimplementing a Bi-LSTM-CNN-Char deep learning architecture on top of Apache Spark, we present a single trainable NER model that obtains new state-of-the-art results on seven public biomedical benchmarks without using heavy contextual embeddings like BERT. 
This includes improving BC4CHEMD to 93.72\% (4.1\% gain), Species800 to 80.91\% (4.6\% gain), and JNLPBA to 81.29\% (5.2\% gain). In addition, this model is freely available within a production-grade code base as part of the open-source Spark NLP library; can scale up for training and inference in any Spark cluster; has GPU support and libraries for popular programming languages such as Python, R, Scala and Java; and can be extended to support other human languages with no code changes.
\end{abstract}


%
\IEEEpeerreviewmaketitle

\section{Introduction}
\label{sec:introduction}

Electronic health records (EHRs) are the primary source of information for clinicians tracking the care of their patients. Information fed into these systems may be found in structured fields for which values are inputted electronically (e.g. laboratory test orders or results)~\cite{liede2015validation} but most of the time information in these records is unstructured making it largely
inaccessible for statistical analysis~\cite{murdoch2013inevitable}. These records include information such as the reason for administering drugs, previous disorders of the patient or the outcome of past treatments, and they are the largest source of empirical data in biomedical research, allowing for major scientific findings in highly relevant disorders such as cancer and Alzheimer’s disease ~\cite{perera2014factors}. Unlocking this information can bring a significant advancement to biomedical research.

The widespread adoption of EHRs and the growing wealth of digitized information sources about patients are opening new doors to uncover previously unidentified associations and accelerating knowledge discovery via state-of-the-art Machine Learning (ML) algorithms and new statistical methods. Due to innate obstacles in extracting information from unstructured text data and the high level of preciseness dictated in healthcare domain, manual abstraction has been prevalent in the industry. As the manual abstraction is highly expensive, time consuming and error prone process, there has been a growing trend in natural language processing (NLP) applications in clinical and biomedical domain to automate the abstraction process as well as making the EHR data available through high-performant and fail-safe pipelines. 

As the  key ingredient of any NLP system, named entity recognition (NER) is regarded as the first building block of question answering, topic modelling, information retrieval, etc~\cite{yadav2019survey}.  In the medical domain, NER plays the most crucial role by giving out the first meaningful chunks of a clinical note, and then feeding them as an input to the subsequent downstream tasks such as clinical assertion status~\cite{uzuner20112010}, clinical entity resolvers ~\cite{tzitzivacos2007international} and de-identification of the sensitive data~\cite{uzuner2007evaluating}. However, segmentation of clinical and drug entities is considered to be a difficult task in biomedical NER systems because of complex orthographic structures of named entities ~\cite{liu2015effects}.

ML methods formulate the clinical NER task as a sequence labeling problem that aims to find the best label sequence (e.g., BIO format labels) for a given input sequence (individual words from clinical text)~\cite{wu2017clinical}.  Many top-ranked NER systems applied the Conditional Random Fields (CRFs) model~\cite{lafferty2001conditional}, which is the most popular solution among conventional ML algorithms. A typical state-of-the-art clinical NER system usually utilizes features from different linguistic levels, including orthographic information (e.g., capitalization of letters, prefix and suffix), syntactic information (e.g. POS tags), word n-grams, word embeddings, and semantic information (e.g., the UMLS concept unique identifier)~\cite{wu2017clinical}. These features are usually utilized in LSTM~\cite{hochreiter1997long} based neural network frameworks~\cite{huang2015bidirectional, chiu2016named, ma2016end} and gained popularity among researchers due to their effectiveness of modeling the sequential patterns. 

In the last few months, pretraining large neural language models and rich contextual embeddings, such as BERT~\cite{devlin2018bert} and ELMO~\cite{peters2018deep}, have also led to impressive gains on NER systems and many clinical variants of BERT models such as BioBert~\cite{lee1901so}, ClinicalBert~\cite{alsentzer2019publicly}, BlueBert~\cite{peng2019transfer}, SciBert~\cite{beltagy2019scibert} and PubmedBert~\cite{gu2020domain} have been crafted to address biomedical and clinical NER tasks with state-of-the-art results. However, since these methods require significant computational resources during both pretraining and getting prediction, using them in production is impractical under the restricted computational resources compared to classical pretrained embeddings (e.g. Glove). A recent study~\cite{arora2020contextual} empirically shows that classical pretrained embeddings can match contextual embeddings on industry-scale data, and often perform within 5 to 10\% accuracy (absolute) on benchmark tasks.

Despite the growing interest and all these ground breaking advances in NER systems, easy to use production ready models and tools are scarce and it is one of the major obstacles for clinical NLP researchers to implement the latest algorithms into their workflow and start using immediately. On the other hand, NLP tool kits specialized for processing biomedical and clinical text, such as MetaMap~\cite{aronson2010overview} and cTAKES~\cite{savova2010mayo} typically do not make use of new research innovations such as word representations or neural networks discussed above, hence producing less accurate results~\cite{zhang2020biomedical, neumann2019scispacy}. In the last year, two new libraries, Stanza~\cite{zhang2020biomedical} and SciSpacy~\cite{neumann2019scispacy} took the stage to find a solution to the issues discussed above and released Python-based, de facto language of data science, production grade libraries. Both libraries offer out of the box clinical and biomedical pretrained NER models utilizing state-of-the-art deep learning frameworks mentioned above. However, none of these libraries or tools can scale up in clusters in terms of distributed data processing principles and do not support in-memory distributed data processing solutions such as Spark.

In this study, we show through extensive experiments that our NER module in Spark NLP library, one of the most widely used NLP libraries in industry, exceeds the biomedical NER benchmarks reported by Stanza in 7 out of 8 benchmark datasets and in every dataset reported by SciSpacy. Using the modified version of the well known BiLSTM-CNN-Char NER architecture~\cite{chiu2016named} into Spark environment, Spark NLP's NER module can also be extended to other spoken languages with zero code changes and can scale up in Spark clusters. 

The specific novel contributions of this paper are the following:
\begin{itemize}

\item Delivering the first production-grade scalable NER model implementation.

\item Delivering a state-of-the-art NER model that exceeds the biomedical NER benchmarks reported by Stanza and SciSpaCy.

\item Comparing the effectiveness of domain specific clinical word embeddings with general purpose GloVe embeddings inside the same NER architecture.

\item Explaining the NER model implementation in Spark NLP which is the only NLP library that can scale up in Spark clusters while supporting popular programming languages (Python, R, Scala and Java).

\end{itemize}

The remainder of the paper is organized as follows: 
Section~\ref{sec:NerDL} introduces Spark NLP and explains the NER model framework implemented in Spark NLP.
Section~\ref{sec:experiments} elaborates the implementation details, datasets and settings for our experiments and presents results for Spark NLP, Stanza and SciSpacy on the same benchmark datasets. 
Section~\ref{sec:conclusion}  concludes this paper by pointing out key points and future directions.

\section{NER Model Implementation in Spark NLP}
\label{sec:NerDL}

The deep neural network architecture for NER model in Spark NLP is BiLSTM-CNN-Char framework, a slightly modified version of the architecture proposed by Chiu et.al.~\cite{chiu2016named}. It is a neural network architecture that automatically detects word and character-level features using a hybrid bidirectional LSTM and CNN architecture, eliminating the need for most feature engineering steps. 

\begin{figure}[h]
\centering
\includegraphics[width=0.4\textwidth,scale=0.4]{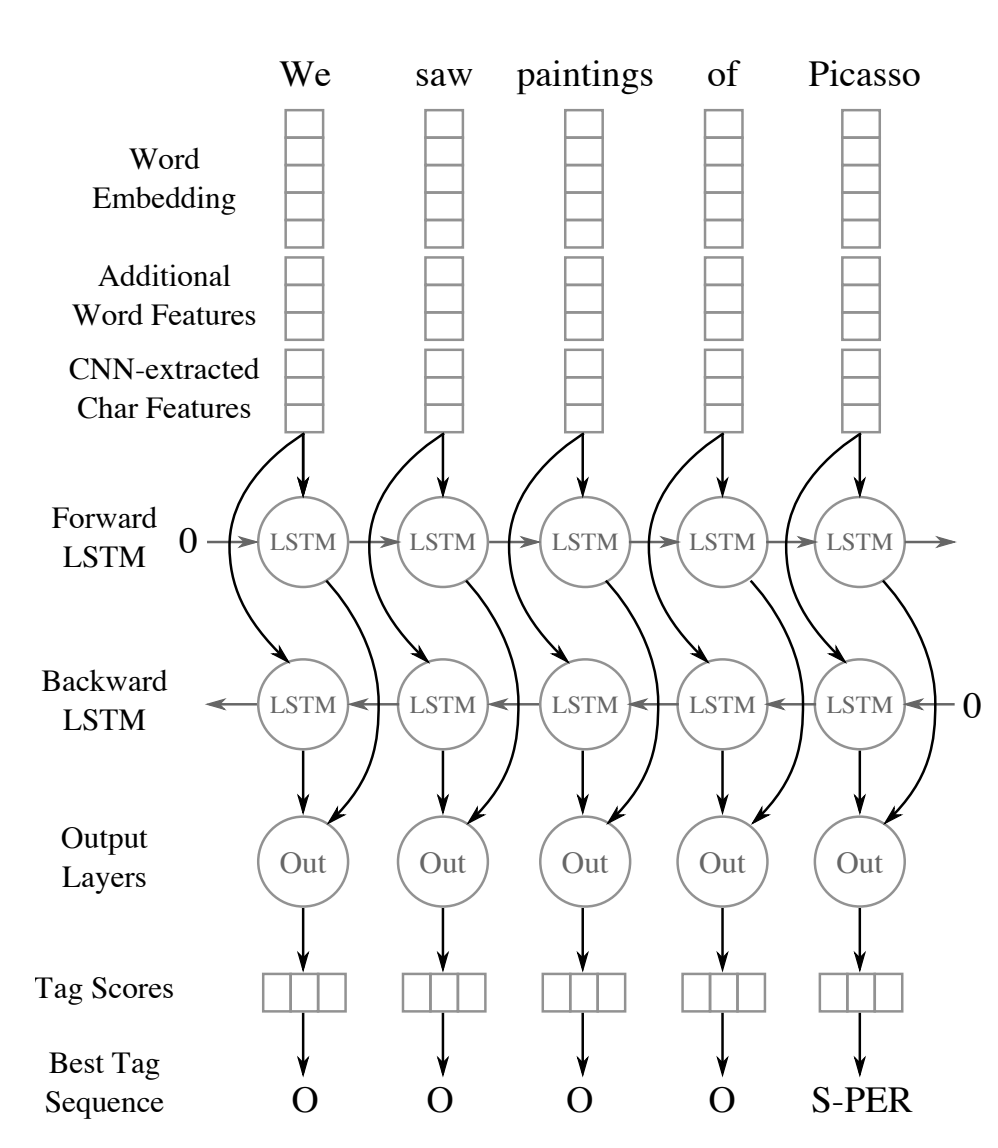}
\centering
\caption{Overview of the original BiLSTM-CNN-Char architecture~\cite{chiu2016named}.}
\label{fig:ner_dl_diagram}
\end{figure}

In the original framework, the CNN extracts a fixed length feature vector from character-level features. For each word, these vectors are concatenated and fed to the BLSTM network and then to the output layers. They employed a stacked bi-directional recurrent neural network with long short-term memory units to transform word features into named entity tag scores. The extracted features of each word are fed into a forward LSTM network and a backward LSTM network. The output of each network at each time step is decoded by a linear layer and a log-softmax layer into log-probabilities for each tag category. These two vectors are then simply added together to produce the final output~\cite{chiu2016named}. The detailed architecture of the proposed framework in the original paper is illustrated at Figure~\ref{fig:ner_dl_diagram}. In sum, 50-dimensional pretrained word embeddings is used for word features, 25-dimension character embeddings is used for char features, and capitalization features (\textit{allCaps, upperInitial, lowercase, mixedCaps, noinfo}) are used for case features. They also made use of lexicons as a form of external knowledge as proposed in ~\cite{ratinov2009design}.

In Spark NLP, we modified this framework as follows:

\begin{itemize}
\item ~\cite{habibi2017deep} compared the performance of LSTM-CRF approach on 33 data sets covering five different entity classes with that of best-of-class NER tools and an entity-agnostic CRF implementation. On average, F1-score of LSTM-CRF is 5\% above that of the baselines, using WikiPubMed-PMC word embeddings.

Using a similar neural network architecture, we trained our own biomedical word embeddings with skip-gram model on PubMed abstracts and case studies, as described in ~\cite{mikolov2013efficient}, for learning distributed representations of words using contextual information. The trained word embeddings has 200-dimensions and a vocabulary size of 2.2 million. In order to compare the effectiveness of this embeddings, we also used 300-dimension pretrained GloVe embeddings with 6 billion tokens, trained on Wikipedia and Gigaword-5 dataset~\cite{pennington2014glove}. Both embeddings are ported into Spark through an annotator concept specifically designed for Spark NLP. The average word coverage of our implementation of domain specific word embeddings (we call it Spark-Biomedical embeddings in this study) is 99.5\% and the coverage of Glove6B embeddings is 96.1\% on the biomedical datasets used in this study (see Table~\ref{tab:oov}).

\item Even though better results were reported by ~\cite{ghaddar2018robust} through robust lexical features, after experimenting with different parameters and components, we decided to remove lexical features in order to reduce the complexity and relied on pretrained biomedical embeddings, casing features and char features through CNN. As sentences are represented through 2 nested sequences (words \& chars), a CNN is applied in a way that each character is embedded in a character embedding matrix, of dimension 25. Then, a 1D Convolution layer processes the sequence of embedded char vectors, followed by a MaxPooling operation. This way, each word gets a vector representation. We used 25 filters and kernel size of 3. It is worth to mention that char features are proved to be highly useful in NER models and had provided a level of immunity to typos and spelling errors. 

\item We built a modified version of the framework~\cite{chiu2016named} in Tensorflow (TF) and used LSTMBlockFusedCell. This is an extremely efficient LSTM implementation based on ~\cite{zaremba2014recurrent}, that uses a single TF operation for the entire LSTM. Our experiments show that it is both faster and more memory-efficient than LSTMBlockCell. Then we implemented this framework in Scala using TensorFlow API. This setup is ported into Spark and let the driver node run the entire training using all the available cores on the driver node. We also added CuDA version of each TF component to be able to train our models on GPU when available.

\end{itemize}

\begin{table*}[ht!]
\caption{Word embeddings coverage ratios on biomedical datasets. Our domain specific embeddings have near-perfect word coverages.  The average word coverage of our implementation of domain specific word embeddings (we call it Spark-Biomedical Embeddings in this study) is 99.5\% and the average word coverage of Glove6B embeddings is 96.1\% on the biomedical datasets used in this study)}
\centering
\bigskip
\label{tab:oov}
\resizebox{0.90\textwidth}{!}{
\begin{tabular}{lLLLL}
\toprule
\multirow{2}{*}{Dataset} & \multicolumn{2}{c}{Spark-Biomedical Embeddings}        & \multicolumn{2}{c}{Spark-Glove6B Embeddings}           \\
                         & Training set & Test set & Training set & Test set\\\midrule
NBCI-Disease             & 99.700                      & 99.695                   & 96.703                    & 96.710                    \\
BC5CDR                   & 99.171                    & 99.106                   & 96.059                    & 95.795                   \\
BC4CHEMD                 & 99.571                    & 99.551                   & 96.409                    & 96.434                   \\
Linnaeus                 & 99.162                    & 99.181                   & 96.801                    & 96.867                   \\
Species800               & 99.350                     & 99.345                   & 95.909                    & 96.258                   \\
JNLPBA                   & 99.530                     & 99.496                   & 92.566                    & 92.690                    \\
AnatEM                     & 99.580                     & 99.623                   & 96.992                    & 96.945                   \\
BioNLP-CG                & 99.859                    & 99.814                   & 97.750                     & 96.663        \\\bottomrule          
\end{tabular}
}
\end{table*}

Due to architectural design choices by Tensorflow implementation in JVM at the time of writing this paper, distributing the model training over the worker nodes in the cluster was not viable and effective, and putting the burden of entire training process on the driver node mandated some limitations in terms of training speed and computational resources. Nevertheless, being able to get predictions on scale from voluminous data with state-of-the-art accuracy would overwhelm the aforementioned disadvantage. 
\section{Implementation Details and Experimental Results}
\label{sec:experiments}

In this section, we describe the datasets, evaluation metrics, and provide an overview of experimental setup.

\subsection{Datasets}

In this study, we trained individual NER models on 8 publicly available biomedical NER datasets provided by ~\cite{wang2019cross}: AnatEM ~\cite{pyysalo2014anatomical}, BC5CDR ~\cite{li2016biocreative}, BC4CHEMD ~\cite{krallinger2015chemdner}, BioNLP13CG ~\cite{pyysalo2015overview}, JNLPBA ~\cite{kim2004introduction}, Linnaeus ~\cite{gerner2010linnaeus}, NCBI-Disease ~\cite{dougan2014ncbi} and S800 ~\cite{pafilis2013species}. These models cover a wide variety of entity types in domains ranging from anatomical analysis to genetics and cellular biology. For the sake of brevity, we didn't include details about the nature of the data sets and readers can refer to cited papers for more information. We trained several other clinical and biomedical NER models in Spark NLP, but we just report metrics on these 8 biomedical data sets as Stanza and SciSpacy also reported their benchmarks on these data sets that are freely available without any restrictions.

\subsection{Overview of Experimental Setup}

Biomedical NER datasets provided by ~\cite{wang2019cross} are already in BIO and BIOES schemes for encoding entity annotations as token tags. IOB (or BIO) stands for Begin, Inside and Outside. Words tagged with O are outside of named entities and the I-XXX tag is used for words inside a named entity of type XXX. Whenever two entities of type XXX are immediately next to each other, the first word of the second entity will be tagged B-XXX to highlight that it starts another entity. On the other hand, BIOES (also known as BIOLU) is a little bit sophisticated annotation method that distinguishes between the end of a named entity and single entities. BIOES stands for Begin, Inside, Outside, End, Single. In this scheme, for example, a word describing a gene entity is tagged with “B-Gene” if it is at the beginning of the entity, “I-Gene” if it is in the middle of the entity, and “E-Gene” if it is at the end of the entity. Single-word gene entities are tagged with “S-Gene”. All other words not describing entities of interest are tagged as ‘O’. 

BIOES scheme was also used in the original implementation of our NER architecture and considerable performance improvements over BIO are reported ~\cite{chiu2016named}. ~\cite{ratinov2009design} also showed that the minimal BIO scheme was more difficult to learn than the BIOES scheme, which explicitly marks boundary tokens. However, we experienced various performance issues when we used BIOES schema (converging very fast in the early epochs but then fail to generalize further and stuck at local minima), and then decided to use BIO scheme. 

In terms of hyperparameter tuning, we run experiments by tuning the hyperparamaters with the following parameter ranges through Random Search ~\cite{bergstra2012random} and found out that the following parameters would produce the best results (figures within the parenthesis represent the parameter ranges tested):
\begin{itemize}
\item LSTM state size: 200 (200, 250)
\item Dropout rate: 0.5 (0.3, 0.7)
\item Batch size: 8 (4, 256)
\item Learning rate: 0.001 (0.01, 0.0003)
\item Epoch: 10-15 (10, 100)
\item Optimizer: Adam
\item Learning rate decay coefficient (po) (\textit{real learning rate = lr / (1 + po * epoch}) ~\cite{smith2018disciplined} : 0.005 (0.001, 0.01))
\end{itemize}

\subsection{Experiment Results}
\begin{figure*}[!ht]
\includegraphics[width=\textwidth]{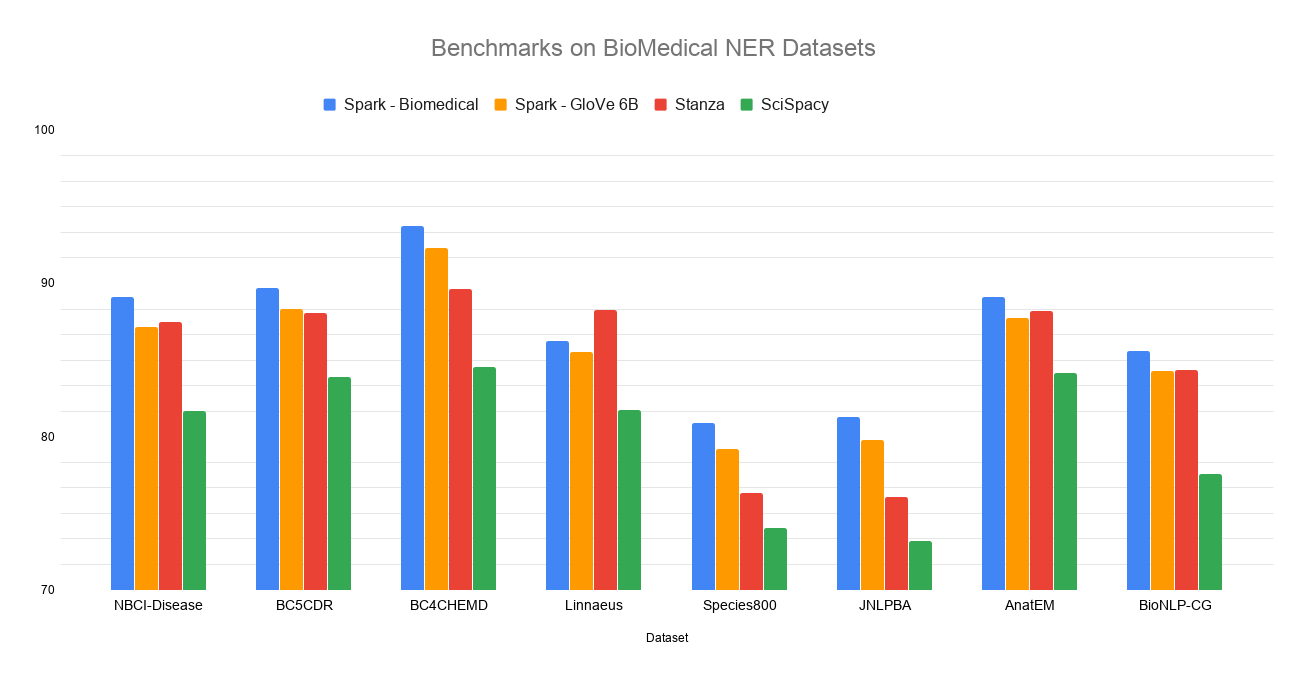}
\caption{NER performance across different biomedical benchmark datasets. Our implementation of NER model with domain specific embeddings exceeds Stanza in 7 out of 8 datasets and exceeds SciSpacy in all the benchmarks. The same implementation with general purpose GloVe embeddings is also better than SciSpacy in every dataset and exceeds Stanza in 4 out of 8 datasets.}
\label{fig:benchmark_bars}
\end{figure*}

We run our experiments on Colab\footnote{https://colab.research.google.com/} server provided by Google (2vCPU @ 2.2GHz, 13GB RAM) and used Apache Spark in local mode (no cluster). We present our results at Table~\ref{tab:benchmarks} and Figure~\ref{fig:benchmark_bars}. As the only NLP library that scales up for training and inference in any Spark cluster, Spark NLP NER architecture obtains new state-of-the-art results on seven public biomedical benchmarks without using heavy contextual embeddings like BERT. This includes improving BC4CHEMD to 93.72\% (4.1\% gain), Species800 to 80.91\% (4.6\% gain), and JNLPBA to 81.29\% (5.2\% gain). Given that Stanza already claims that its NER performance is on par with or superior to the strong performance achieved by BioBERT, our proposed NER model can get better results despite using considerably more compact model. Moreover, this model is available within a production-grade code base as part of the open-source Spark NLP library and a new NER model can be trained with a single line of code as presented in Appendix~\ref{sec:appendix}.

\begin{table*}[!ht]
\caption{NER performance across different datasets in the biomedical domain. All scores reported
are micro-averaged test F1 excluding O's. Stanza results are from the paper reported in ~\cite{zhang2020biomedical}, SciSpaCy results are from the scispacy-medium models reported in ~\cite{neumann2019scispacy}. The official training and validation sets are merged and used for training and then the models are evaluated on the original test sets.  For reproducibility purposes, we use the preprocessed versions of these datasets provided by ~\cite{wang2019cross} and also used by Stanza. Spark-x prefix in the table indicates our implementation. Bold scores represent the best scores in the respective row.}
\centering
\label{tab:benchmarks}
\resizebox{0.97\textwidth}{!}{
\begin{tabular}{llLLll}
\toprule
Dataset  & Entities   & Spark - Biomedical & Spark - GloVe 6B & Stanza & SciSpacy \\ \midrule
NBCI-Disease & Disease                                 & \textbf{89.13}                       & 87.19                     & 87.49  & 81.65    \\
                     BC5CDR       & Chemical, Disease                       & \textbf{89.73}                       & 88.32                     & 88.08  & 83.92    \\
                     BC4CHEMD     & Chemical                                & \textbf{93.72}                       & 92.32                     & 89.65  & 84.55    \\
                     Linnaeus     & Species                                 & 86.26                       & 85.51                     & \textbf{88.27}  & 81.74    \\
                     Species800   & Species                                 & \textbf{80.91}                       & 79.22                     & 76.35  & 74.06    \\
                     JNLPBA       & 5 types in cellular & \textbf{81.29}                       & 79.78                     & 76.09  & 73.21    \\
                     AnatEM       & Anatomy                                 & \textbf{89.13}                       & 87.74                     & 88.18  & 84.14    \\
                     BioNLP13-CG  & 16 types in Cancer Genetics & \textbf{85.58}  & 84.3 & 84.34  & 77.6     \\ \bottomrule
\end{tabular}
}
\end{table*}

As you can see on the leaderboard given at Table~\ref{tab:leaderboard}, our NER model with pretrained biomedical embeddings produces better results than Stanza in 7 out of 8 biomedical datasets and exceeds SciSpacy in all the benchmarks. It is also surprising to see that our NER model with GloVe6B embeddings, despite being a general purpose embeddings, can also exceed Stanza's (also using domain specific embeddings, CharLM - character-level language model~\cite{akbik2018contextual}) benchmarks in half of the benchmarks and again exceeds SciSpacy in all the benchmarks.

\begin{table*}[ht!]
\caption{Biomedical NER benchmarks leaderboard. Spark-x prefix indicates our implementation.}
\centering
\bigskip
\label{tab:leaderboard}
\resizebox{0.9\textwidth}{!}{
\begin{tabular}{lllllll}
\toprule
\multirow{3}{*}{Dataset} & \multicolumn{2}{c}{Best} & \multicolumn{2}{c}{2nd Best} & \multicolumn{2}{c}{3rd Best} \\ 
                         & Model           & Score  & Model             & Score    & Model            & Score     \\\midrule
NBCI-Disease             & Spark-Biomedical  & 89.13  & Stanza            & 87.49    & Spark-GloVe6B      & 87.19     \\
BC5CDR                   & Spark-Biomedical  & 89.73  & Spark-GloVe6B       & 88.32    & Stanza           & 88.08     \\
BC4CHEMD                 & Spark-Biomedical  & 93.72  & Spark-GloVe6B       & 92.32    & Stanza           & 89.65     \\
Linnaeus                 & Stanza          & 88.27  & Spark-Biomedical    & 86.26    & Spark-GloVe6B      & 85.51     \\
Species800               & Spark-Biomedical  & 81.29  & Spark-GloVe6B       & 79.78    & Stanza           & 76.09     \\
AnatEM                     & Spark-Biomedical  & 89.13  & Stanza            & 88.18    & Spark-GloVe6B      & 87.74     \\
BioNLP-CG                & Spark-Biomedical  & 85.58  & Stanza            & 84.34    & Spark-GloVe6B      & 84.3     \\ \bottomrule
\end{tabular}
}
\end{table*}

\section{Conclusion}
\label{sec:conclusion}

Despite the growing interest and ground breaking advances in NLP research and NER systems, easy to use production ready models and tools are scarce in Biomedical domain and it is one of the major obstacles for clinical NLP researchers to implement the latest algorithms into their workflow and start using immediately. 

In this study, we show through extensive experiments that NER module in Spark NLP library, one of the most widely used NLP libraries in industry, exceeds the biomedical NER benchmarks reported by Stanza in 7 out of 8 benchmark datasets and in every dataset reported by SciSpacy without using heavy contextual embeddings like BERT. Using the modified version of the well known BiLSTM-CNN-Char NER architecture~\cite{chiu2016named} into Spark environment, we also presented that even with a general purpose GloVe embeddings (GloVe6B) and with no lexical features, we were able to achieve state-of-the-art results in biomedical domain and produces better results than Stanza in 4 out of 8 benchmark datasets. Given that Stanza also uses domain specific clinical embeddings, exceeding its benchmarks with general purpose embeddings is also another important observation.

Spark NLP's NER module can also be extended to other spoken languages with zero code changes and can scale up in Spark clusters. In addition, this model is available within a production-grade code base as part of the Spark NLP library; can scale up for training and inference in any Spark cluster;  has GPU support and libraries for popular programming languages such as Python, R, Scala and Java; and is already extended to support other human languages with no code changes. 

\bibliographystyle{unsrtnat}
\bibliography{References}


\appendix

\section{Appendices}
\label{sec:appendix}

\begin{lstlisting}[language=Python]

from pyspark.ml import Pipeline
import sparknlp
from sparknlp.training import CoNLL
from sparknlp.annotator import *

spark = sparknlp.start()

training_data = CoNLL().readDataset(spark, 'BC5CDR_train.conll')

word_embedder = WordEmbeddings.pretrained('wikiner_6B_300', 'xx') \
 .setInputCols(["sentence",'token'])\
 .setOutputCol("embeddings")

nerTagger = NerDLApproach()\
  .setInputCols(["sentence", "token", "embeddings"])\
  .setLabelColumn("label")\
  .setOutputCol("ner")\
  .setMaxEpochs(10)\
  .setDropout(0.5)\
  .setLr(0.001)\
  .setPo(0.005)\
  .setBatchSize(8)\
  .setValidationSplit(0.2)\

pipeline = Pipeline(
    stages = [
    word_embedder,
    nerTagger
  ])

ner_model = pipeline.fit(training_data)
\end{lstlisting}

\end{document}